\documentclass[twocolumn, switch]{article} % Method A for two-column formatting

\usepackage{preprint}

\usepackage{amsmath, amsthm, amssymb, amsfonts}
 \usepackage{relsize}
 \usepackage{multirow}

\usepackage[numbers,square]{natbib}
\usepackage[utf8]{inputenc}	% allow utf-8 input
\usepackage[T1]{fontenc}	% use 8-bit T1 fonts
\usepackage{booktabs} 		% professional-quality tables
\usepackage{nicefrac}		% compact symbols for 1/2, etc.
\usepackage{microtype}		% microtypography
\usepackage{lineno}		% Line numbers
\usepackage{float}			% Allows for figures within multicol
\usepackage{titlesec}
\titlespacing\section{0pt}{12pt plus 3pt minus 3pt}{1pt plus 1pt minus 1pt}
\titlespacing\subsection{0pt}{10pt plus 3pt minus 3pt}{1pt plus 1pt minus 1pt}
\titlespacing\subsubsection{0pt}{8pt plus 3pt minus 3pt}{1pt plus 1pt minus 1pt}

\usepackage{hyperref}

\title{NODE-SELECT: A Graph Neural Network Based On A Selective Propagation Technique}

\usepackage{authblk}

\author[1]{Steph-Yves Louis}
\author[1]{Alireza Nasiri}
\author[2]{Fatima J. Rolland}
\author[3]{Cameron Mitro}
\author[1]{Jianjun Hu}

\affil[1]{Department of Computer Science \& Engineering, University of South Carolina}
\affil[2]{Department of Applied Behavioral Analysis, Drexel University}
\affil[3]{School of Medicine, University of South Carolina}

\begin{document}

\twocolumn[ % Method A for two-column formatting
  \begin{@twocolumnfalse} % Method A for two-column formatting
  
\maketitle

\begin{abstract}
While there exists a wide variety of graph neural networks (GNN) for node classification, only a minority of them adopt mechanisms that effectively target noise propagation during the message-passing procedure. Additionally, a very important challenge that significantly affects graph neural networks is the  issue of scalability which limits their application to larger graphs. In this paper we propose our  method named NODE-SELECT: an efficient graph neural network that uses subsetting layers which only allow the best sharing-fitting nodes to propagate their information. By having a selection mechanism within each layer which we stack in parallel, our proposed method NODE-SELECT is able to both reduce the amount noise propagated and adapt the restrictive sharing concept observed in real world graphs. Our NODE-SELECT significantly outperformed  existing GNN frameworks in noise experiments and matched state-of-the art results in experiments without noise over different benchmark datasets.
\end{abstract}
 % (optional)
\vspace{0.35cm}

  \end{@twocolumnfalse} % Method A for two-column formatting
] % Method A for two-column formatting

%\begin{multicols}{2} % Method B for two-column formatting (doesn't play well with line numbers), comment out if using method A

%%%%%%%%%%%%%%%  Main text   %%%%%%%%%%%%%%%
% \linenumbers

\section{Introduction}
The use of deep learning techniques for graph analysis has become a very popular research topic in recent years \cite{zhou2018graph}. Commonly referred to as graph neural networks (GNN), these deep learning techniques now figure amongst the most used methods for learning from relational data \cite{zhou2018graph,wu2020comprehensive}. Just as in the functioning of convolutional neural networks (CNN), multiple convolution operations can also be applied to learn from non-Euclidean data \cite{zhou2018graph,wu2020comprehensive,lecun2015deep}. Various adaptations of GNNs have been proposed over the years for the purpose of node classification \cite{zhou2018graph,wu2020comprehensive}. These GNN adaptations mainly differ in regards to their node embedding techniques, their algorithms' propagation or aggregation methods, and their model's scalability \cite{zhou2018graph,wu2020comprehensive}. Examples of important GNN variants include GCN \cite{kipf2016semi}, GAT \cite{velivckovic2017graph}, GraphSAGE \cite{hamilton2017inductive}, DeepGCN \cite{li2019deepgcns}, and Pairnorm as one of the most recent works \cite{zhao2019pairnorm}. 

Prior GNNs have made use of regularization, Chebyshev polynomials, feature normalization, attention, residual blocks, random sampling, and many other techniques which have further pushed state-of-the-art for GNN \cite{zhou2018graph,wu2020comprehensive,kipf2016semi,velivckovic2017graph,hamilton2017inductive,li2019deepgcns,rong2019dropedge,fey2019just}. Nevertheless, there still remain many factors that still present challenges to GNN \cite{zhou2018graph,wu2020comprehensive,kipf2016semi}. Two important factors that mostly get referenced in the literature are over-smoothing and overfitting.  Over-smoothing, defined as excessive similarity of node representation, is a direct consequence of deeply stacking graph convolutional layers \cite{li2018deeper,zhou2018graph,chen2020measuring}. Particularly, the over-smoothing issue has been justified to result from the fact that more noise gets shared than useful information during the convolution operations \cite{chen2020measuring}. On the other hand, overfitting is another issue which happens with adding more parameters and increasing the complexity of the model \cite{rong2019dropedge}. Besides the over-smoothing and over-fitting issues, GNN also suffer from a noticeable conceptual limitation. Few GNN variants provide an implementation that fully mimic the relational rules observed in the networks of real world \cite{velivckovic2017graph,zhang2018gaan,rong2019dropedge,zhao2019pairnorm,louis2020graph,liu2020towards}. Nonetheless, the number of GNN variants with mechanisms that easily translate to real-world networks is minimal and remains to be further exploited; particularly, scenarios in which there are consequences to letting any nodes propagate information \cite{shchur2018pitfalls,leo2019many,rese2013too,feily2009survey}. 

With the motivation to mainly tackle this existing conceptual limitation, we propose a new kind of graph neural network named: NODE-SELECT \footnote{The codes for this work can be found at \url{https://github.com/superlouis/NODE-SELECT}}. For this conceptual limitation, we introduce an efficient selection mechanism that prevents nodes with representation of poorer quality from propagating information. Beyond the selection process, we also learn a global weight weight coefficient for these propagating nodes and also combine our memory-efficient layers in parallel as in the ensemble concept. To demonstrate the effectiveness and importance of our proposed method, we evaluate it on standard benchmark datasets with and without noise data \cite{sen2008collective,bojchevski2017deep,shchur2018pitfalls}. Overall, our proposed NODE-SELECT considerably outperforms popular GNN frameworks on graphs augmented with noise vertices and marginally surpasses them on graph without noise vertices.

We summarize our contribution as four main points. \textbf{(1)} We implement a very important concept of node selection to graph neural networks. \textbf{(2)} We adapt the ensemble concept by stacking our graph convolutional layers in parallel and demonstrate how it benefits our framework. \textbf{(3)} We demonstrate through extensive experiments how current GNN can be considerably affected by the presence of harmful vertices representations during the message-passing but our NODE-SELECT is not affected by them. \textbf{(4)} We demonstrate that our proposed method is extremely scalable with the increase of graph sizes.

\section{RELATED WORK}
\label{related_work}
Researchers have used various techniques to do their convolution operation on the graph vertices.
Examples of such techniques include the adaptation of gated recurrent units (GRU), Chebyshev polynomials, attention mechanisms, etc... GGNN was the first framework to apply gated recurrent units to sequentially  update the feature vectors of the graph nodes \cite{li2015gated,cho2014learning}. While ChebConv first used Chebyshev polynomials to do the node convolutions, the adaptation by GCN proved to be more effective thanks to its feature aggregation restriction and normalization trick \cite{defferrard2016convolutional,kipf2016semi}. The  technical concept of sampling was  introduced by the works of GraphSAGE and FastGCN \cite{hamilton2017inductive,fey2019fast}. Another important technical concept: attention, for graph neural networks, was first adapted in the framework of GAT \cite{velivckovic2017graph}. 

In addition to the aforementioned methods, there exist many GNN variants that have further incrementally introduced other important techniques into the field of graph neural networks. Such architectures include DropEdge and DNA-Conv \cite{rong2019dropedge,xu2018representation}. DropEdge proposes a regularization mechanism to address over-fitting and over-smoothing by randomly removing connecting edges \cite{rong2019dropedge}. On the other hand, inspired by the concepts of Jumping Knowledge \cite{xu2018representation}, Fey proposed a dynamic neighborhood aggregation mechanism to offer to their learning model a bigger range of feature information \cite{fey2019just}. Nevertheless, there still remains conceptual limitations that still need to be addressed to further advance the field. 

The main limitation we aim to address is the need for \textit{a straight-forward adaptation of the natural message-passing mechanism} often found in real-world graphs. In real graphs such as social networks, computer networks, brains, or molecules; we frequently observe an orderly communication between the units. Our goal is to implement this communication of real-world graphs in which only a subset of the vertices actively exchange information simultaneously. Works in the fields of Sociology and Neuroscience have studied examples of this orderly communication. In Sociology, numerous publications have studied the relatable topic that only the best suitable people should lead tasks within a social network \cite{leo2019many,rese2013too,groysberg2011too}. This restriction of propagating vertices in a social setting is often paraphrased as 'Too many cooks spoil the broth'.
In Neuroscience there has also been works that studied the topic that only a subset of neurons fire simultaneously in brain networks \cite{havenith2011synchrony,sasaki2014interneuron}. Just as there is a clear limitation in propagating vertices in some real-world graphs, we also implement in our NODE-SELECT a similar mechanism that grants  our model the flexibility to adapt this restriction. 

Besides the need of adapting this conceptual selection mechanism, we also assumed that the implementation of such  mechanism would also technically benefit the network. This adapted selection mechanism could act as a regularization in the network while preventing the least sharing-fit nodes from propagating their embeddings \cite{oymak2018learning,fox2019robust}. This restriction would mainly reduce the amount of noise coming from particular nodes.  With the cancellation of \textit{inappropriate nodes'} propagation, the network would also benefit in efficiency having to arguably do less node convolutions \cite{defferrard2016convolutional,zhou2018graph}. With the selection implemented within each layer, we also presumed that ensembling our layers could be more beneficial than sequentially them. The layers' sequential stacking could lead to poorer performance if prior selections were sub-optimal. Also, ensembling the layers could result in a diverse generation of embeddings which would likely increase accuracy performance of the model \cite{jan2018optimizing}.

% add this paper to related work --> https://arxiv.org/pdf/1912.10206.pdf
% spend some time reading about this --> https://openreview.net/forum?id=H38f_9b90BO

\section{Proposed Method}
\label{nsgnn_method}
% In this section, we describe the technical adaptation of the sharing constraint that inspired our method.
We begin by formulating an input graph $\mathcal{G}$ as the set $\mathcal{G} = (\mathcal{V},\mathcal{E})$, where $\mathcal{V}$ and $\mathcal{E}$ respectively define a set of nodes $v_i$ and a set of edges $e_{ij}$ connecting 2 nodes: $v_i$ and  $v_j$. Also, we denote as $\mathcal{X}=\{x_1,...,x_n\}$ or $\{h^{(0)}_1,...,h^{(0)}_n\} \in \mathbb{R}^{N \times F}$ the node features. For the task of node classification, a graph neural network needs to learn an embedding $h^{(l)}_i$ based on a prior embedding or feature vector $h^{(l-1)}_i$. The operation  done by each layer can be expressed by a function $f^{(l)}_{\theta}$:
\begin{equation}
    h_i^{(l)}=f^{(l)}_{\theta}\left(h_i^{(l-1)},\Gamma\left\{h_{j}^{(l-1)}\right\}_{j\in \mathcal{N}_{(i)}}\right)\label{eq:1}
\end{equation}
 where $\theta$ denotes the trainable parameters of layer $l$, $\Gamma$ any function aggregating localized neighborhood information, and $\mathcal{N}_{(i)}$ the neighborhood of node $v_i$.
%  For instance, the $f^{(l)}_{\theta}$ operation  in GCN takes the form of 
%  \begin{equation}
%     f^{(l)}_{\theta} = \hat{A}H^{(l-1)}W^{(l)}
%  \end{equation}
%  , in which $\theta$ is represented by the learnable matrix $W^{(l)}$, $H^{(l-1)}$ the previous embedding and $\Gamma$ represented by  the normalized adjacency matrix $\hat{A}$: $\hat{A}=\hat{D}^{-1/2}(A+I)\hat{D}^{-1/2}$.
 
 To update node $v_i$'s embedding, these layers generally utilize information from all neighboring nodes and need to be arranged in sequence within the network. However, such layout favours the potential issue of oversmoothing and lacks adequate techniques to prevent the propagation of noise from specific nodes. In contrast, we propose to use a selection mechanism $S(\cdot)$ to limit noise sharing between nodes and the combination of embeddings from independent layers to reduce noise propagation between layers.
 
 \subsection{Node Selection}
The first step in our method consists of linearly transforming the initial feature vector $x_i$ using a matrix $\textbf{W} \in \mathbb{R}^{F'\times F}$. The transformed feature vector $\textbf{W}x_i$, with reduced dimensionality, has the same cardinality as the embedding output. Subsequently, we estimate a random sensitivity value $\hat{p}_i$ as a measure to classify nodes with potentially harmful or useless embedding information. The goal is to use this $\hat{p}_i$ value so the network can detect nodes whose omission of information improves the training. We utilize the sum of the neighbours embedding $\mathcal{N}_{(i)}$ to learn this value and then define our selection technique as:

\begin{equation}
        S(v_i)= 
        \left\{ \begin{array}{ll}
            1, & \hat{p}_i \ge T \\
            0, &\text{otherwise}
        \end{array} \right.
    \end{equation}
 
where $\hat{p}_i=\sigma\left(\textbf{W}_0\left(\sum_{j \in \mathcal{N}_{(i)}}{\textbf{W}x_j} \right)\right)$, $\textbf{W}_{0} \in \mathbb{R}^{1\times F'}$ is a weight matrix, $T$ a determined threshold, and $\sigma$ a non-linear transformation. Thus, $\hat{p}_i$ can be interpreted as a normalized signal, ranging between 0 and 1, allowing the model to make its selection with respect to the entire graph. Namely, the learnable choice of whether or not a node needs to propagate its information is made on a global scale such that an un-selected node can no longer affect any other nodes. 

\subsection{Selective Aggregation and Feature Update}
 Once a node's selection $S(v_i)$ is learned, we allow the layer to aggregate the embedding information only from its selected neighbors. In other words, we want the model to keep learning embeddings even with the cancellation of some nodes' propagation. The simplified form of this selectively aggregated information is simplified below:
 \begin{equation}
        A_{i}= \mathlarger{\sum}_{j \in \mathcal{N}_{(i)}}{ \alpha_{j}\cdot\mathcal{S}(v_j)\cdot\textbf{W}x_j}
    \end{equation}
 where $\alpha_i$ defines the propagation weight for each selected node. This propagation weight is calculated by linearly transforming the concatenated summed and selectively aggregated neighborhood embeddings; such that $\alpha_i = \sigma\left (\textbf{W}_1 \left (\sum_{j \in \mathcal{N}_{(i)}}{\mathcal{S}(v_j)\cdot\textbf{W}x_j} \parallel \sum_{j \in \mathcal{N}_{(i)}}{\textbf{W}x_j}\right ) \right)$ and $\textbf{W}_{1} \in \mathbb{R}^{1\times 2\cdot F'}$ is a weight matrix. Since our model learns the selection $\mathcal{S}(v_j)$ with respect to the entire graph (global selection),  $\alpha_{i}$ thereby corresponds to the hard attention for a selected node $v_i$. Also in our experiments, applying this global weight $\alpha_{i}$ resulted in higher ( $\ge$ 3\%) accuracy performance than adapting a weight relative to each node's neighbor (i.e. $\alpha_{ij}$ as in GAT).

Therefore, for layer $l$ in our method, the simplified form of its function $f_{\theta}^{(l)}$ takes the form of:
 \begin{equation}
     f^{(l)}_{\Theta}(x_i) = \sigma(\textbf{W}x_{i}) +  \sigma(A_{i}) \label{eq:2}
 \end{equation}
 
where $\textbf{W}x_{i}$ is the originally transformed feature vector and $A_{i}$ the pooled embedding information from just the selected nodes.  

\subsection{Parallel Stacking}
The last step consists of simply summing the embedding information from all layers. Given $L$ layers, the final embedding output for a given is the summation of all $L$ independently learned embeddings $f^{(l)}_{\Theta}(x_i)$. This summation is described below.
\begin{equation}
    \text{output}_{i} = \sum_{l=1}^{L}{f^{(l)}_{\Theta}(x_i)}
\end{equation}

\begin{figure*}[ht]
%   \vskip 0.2in
  \centering
  \includegraphics[width=0.8\linewidth,height =3in]{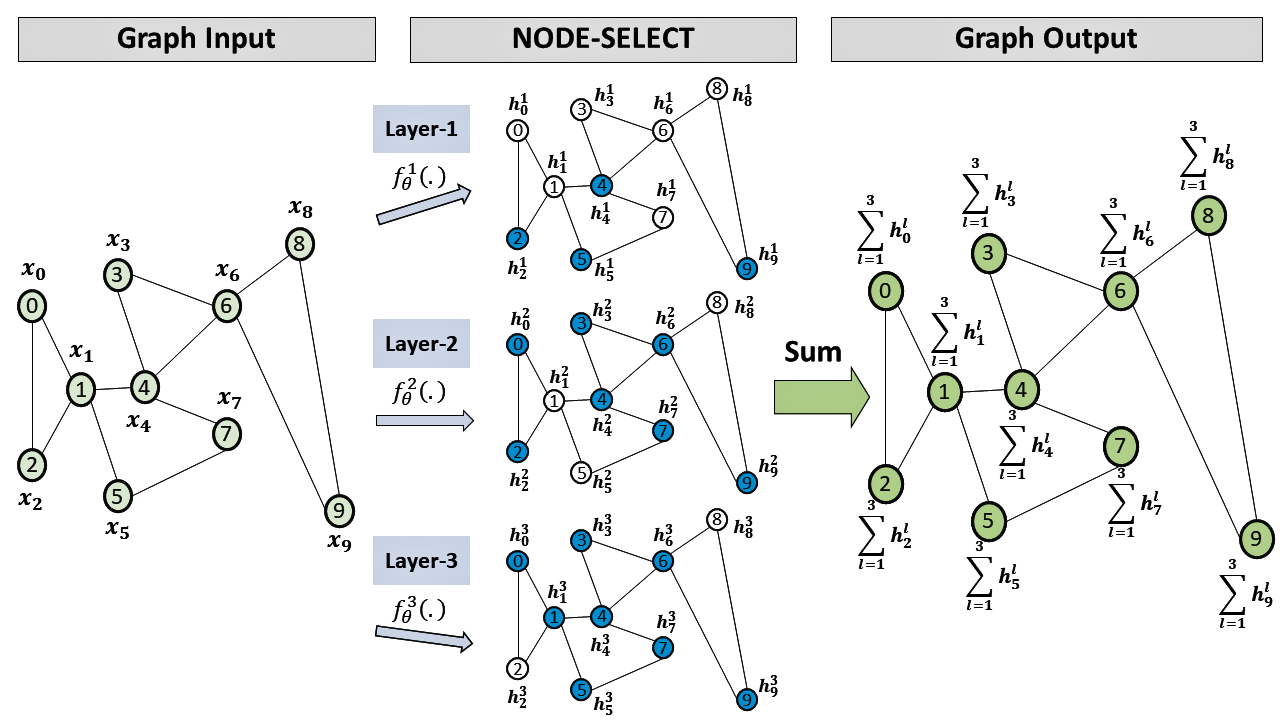}
  \caption{Architecture of the NODE-SELECT graph neural network. Provided a graph, a number of independent layers (3 in this figure) are applied in parallel, where each provides a different state embedding based on their selection of propagating nodes (in blue). Finally, the output of all layers are summed to create the nodes' final hidden state.}
\label{fig:architecture-NS}
\vskip 0.2in
\end{figure*}

\section{EXPERIMENTS}
\subsection{Datasets}
To assess the performance of our proposed model, we conduct two sets of experiments using a total of 8 benchmark datasets. In the first experiment, we utilize all 8 datasets: Cora, CiteSeer, PubMed, Cora Full, Coauthor CS,  Coauthor Physics, Amazon Computers, and  Amazon Photo. Cora, CiteSeer, and PubMed contain relational data on academic papers \cite{sen2008collective, bojchevski2017deep}. Datasets Coauthor CS (Co-CS) and Coauthor Physics (Co-P) are co-authorship datasets from the Microsoft Academic Graph \cite{shchur2018pitfalls}. Lastly, Amazon Computers (Amz-C) and Amazon Photo (Amz-P) are graph datasets defining segments of the Amazon product categories graphs. For each dataset, we randomly split the nodes so that the training, validation, and testing sets follow a ratio of 20-20-60 percent. This split is repeated for 10 randomly chosen seeds which are used in each model experiment. 

In the second experiment, we only use Cora, CiteSeer, and the PubMed datasets which we modify by adding noise data into their graphs. We increase the size of each graph by 10 and 25\% using pseudo vertices. We attribute each pseudo vertex a feature vector from a standard normal distribution, a random label, and random neighbors from the original graph. We follow the same 20-20-60 splitting ratio as in the first experiment but afterwards remove the pseudo vertices from the testing set. This split is repeated for 5 randomly chosen seeds. Details of the Datasets are provided in Table \ref{tab:table_dataset}. 

\begin{table}[ht]
\caption{Statistics of transductive Datasets used in this paper.}
\label{tab:table_dataset}
\begin{center}
% \resizebox{\textwidth}{!}{%
 \begin{tabular}{|c c c c c |}
  \hline
\textbf{Dataset} & \textbf{Nodes} & \textbf{Edges} & \textbf{Classes}& \textbf{Features} \\ [0.3ex] 
 \hline\hline
 CiteSeer & 3,327 & 4,552 & 6 &3,703 \\  \hline
Cora & 2,708 & 5,278 & 7 &1,433 \\ \hline
 PubMed & 19,117 & 44,324& 3 & 500 \\
\hline 
Co-P & 34,493 & 247,962& 5 & 8,415  \\
 \hline 
Co-CS & 18,333 & 81,894& 15 & 6,805  \\
 \hline
Cora Full & 19,793 &  63,421& 70 &  8,710 \\
 \hline 
Amz-P & 7,650 & 245,861& 8 & 8,415 \\
 \hline 
Amz-C & 13,752 & 119,081& 10 & 767  \\
 \hline 
\end{tabular}
\end{center}
\end{table}

\subsection{Experimental Setup}
 We compare our proposed method to 6 GNN variants selected for either their robust performance, contrasting node sampling method, or both. These baselines include DropEdge, FastGCN, GAT, GCN, GraphSAGE, and Node2vec \cite{rong2019dropedge,chen2018fastgcn,velivckovic2017graph,kipf2016semi,hamilton2017inductive,grover2016node2vec}. Given that each framework performs differently under various training dynamics, we perform random hyper-parameter and only report results from the best performing models with respect to the validation set.  We apply a fixed dropout rate of 0.5 after the GNN layers and use Adam as optimizer\cite{kingma2014adam, srivastava2014dropout}. We implement all the models using Pytorch and the library of Pytorch-Geometric \cite{paszke2017automatic,fey2019fast}. The hyper-parameters from the best performing models are provided in Table 1 of the Appendix.

\subsection{Results}
Table \ref{tab:table_results} displays the average accuracies over the 10 random splits from the first experiment. As seen, NODE-SELECT consistently matches or outperforms the performance by the baselines by up to 1.4 percentage points. Table \ref{tab:table_results2} lists the average classification accuracy of 5 random splits for the second experiment with noise introduced to the training. Once a GNN model is introduced to a considerably amount of noise information, the results demonstrate that the accuracy significantly drops \cite{fox2019robust}. Nevertheless, our NODE-SELECT is only marginally affected by the presence of noise information whereas the baselines considerably are. As shown in the results, our proposed method particularly stands out in these noise experiments by outperforming the other baselines by up to 20 percentage points. Simply put, the resilient ability of our network to be affected by noise information is due to the used selection mechanism which allows the network a direct control of blocking nodes propagating them. 

\begin{table*}%[ht]
\caption{Results from the first experiment with the 8 standard benchmarks. Average testing accuracy (\%) and standard deviation from 10 random splits are listed.}
\label{tab:table_results}
\begin{center}
\resizebox{\textwidth}{!}{%
 \begin{tabular}{||c | c | c | c| c | c | c | c| c ||}
  \hline
Variant & CiteSeer & Cora & PubMed & Co-P & Co-CS & Cora Full &Amz-C &Amz-P \\ [0.3ex] 
 \hline\hline
 DropEdge-GCN&56.2$\pm$1.6 &83.5$\pm$2.3 &87.1$\pm$0.5 &95.9$\pm$0.1 &92.8$\pm$0.6 & 57.4$\pm$1.8&84.9$\pm$3.4 & 89.2$\pm$4.1\\
 \hline
FastGCN&74.0$\pm$1.0 &82.1$\pm$2.6&87.6$\pm$0.5&95.5$\pm$0.3 &92.2$\pm$0.4&60.8$\pm$1.0&83.5$\pm$2.2&91.0$\pm$0.9\\
 \hline   
GAT&\textbf{74.2$\pm$0.8}&\textbf{86.0$\pm$0.7}&86.4 $\pm$0.3 &95.7$\pm$0.1 &92.2 $\pm$0.2 &64.8 $\pm$0.5&90.0$\pm$0.7&93.7$\pm$0.6 \\ 
 \hline 
GCN &74.0 $\pm$0.7  &85.0$\pm$0.7 & 87.2$\pm$0.3 &95.9$\pm$0.1 &93.1$\pm$0.2 &\textbf{67.3 $\pm$0.5}&89.4$\pm$0.5&93.5$\pm$0.2\\   \hline
GraphSAGE&73.7 $\pm$0.7&\textbf{86.0$\pm$0.7} & 86.2 $\pm$0.3 &95.4 $\pm$0.2  &93.4 $\pm$0.2 &64.9 $\pm$0.3&\textbf{90.2 $\pm$0.5}&\textbf{94.4 $\pm$0.5}\\
 \hline
Node2vec&55.3$\pm$0.7&78.1$\pm$0.8&80.2 $\pm$0.4 &93.0$\pm$0.1 &87.7$\pm$0.3 &58.8$\pm$0.3&87.2$\pm$0.4&91.0$\pm$0.3\\
 \hline  
NODE-SELECT&74.1$\pm$1.1&\textbf{86.0$\pm$0.7}&\textbf{88.1$\pm$0.3} &\textbf{96.5$\pm$0.1} &\textbf{94.8$\pm$0.1} &\textbf{67.3 $\pm$0.6} &89.6$\pm$0.4&\textbf{94.4$\pm$0.4}\\ 
\hline
\end{tabular}}
\end{center}
\end{table*}

\begin{table*}%[ht]
\caption{Results from the second experiment with the Cora, Citeseer, and Pubmed benchmarks. Graphs sizes from the datasets are increased by 10 and 25\% from the addition of noise data. Average testing accuracy (\%) and standard deviation from 5 random splits are listed.}
\label{tab:table_results2}
\begin{center}
 \begin{tabular}{|c || c | c || c| c || c | c |}
  \hline
\multirow{2}{*}{Variant} & 
\multicolumn{2}{c||}{Citeseer} &
\multicolumn{2}{|c||}{Cora} &
\multicolumn{2}{|c|}{Pubmed} \\\cline{2-7}
 &+10\%  &+25\% & +10\% & +25\% & +10\% & +25\% \\ [0.3ex] 
 \hline\hline
DropEdge-GCN&42.0$\pm$2.3&35.6$\pm$1.4&38.1 $\pm$3.4 &34.4 $\pm$2.9&74.4$\pm$12.0 & 46.2$\pm$12.3\\  \hline
FastGCN&33.7$\pm$0.9&29.9$\pm$1.5&40.8 $\pm$2.1 &30.3$\pm$1.2&57.3$\pm$0.9 & 47.9 $\pm$1.3 \\  \hline
GAT&34.1$\pm$1.2&34.1$\pm$1.8&61.7$\pm$2.0& 58.0$\pm$1.4&55.0 $\pm$4.8 &52.5$\pm$7.5 \\ \hline 
GCN&56.0$\pm$0.8&49.3$\pm$1.1&74.3 $\pm$1.7  &65.2$\pm$1.3 &58.4$\pm$0.6 &54.8$\pm$0.6 \\  \hline
GraphSAGE&35.4$\pm$1.7&33.9$\pm$0.7&52.8$\pm$1.3 &51.9$\pm$1.7 & 45.4$\pm$0.5  &42.7$\pm$1.1  \\ \hline
Node2vec&38.7$\pm$0.8&35.4$\pm$0.8&58.4$\pm$1.4 & 51.1$\pm$1.3&62.9$\pm$0.4  & 58.5$\pm$0.3\\  \hline  
NODE-SELECT&\textbf{68.6$\pm$0.4}&\textbf{64.8$\pm$2.0}&\textbf{80.9$\pm$3.8}&\textbf{78.4$\pm$1.8}&\textbf{83.6$\pm$0.9} &\textbf{79.7$\pm$0.8 }\\ \hline
\end{tabular}
\end{center}
\end{table*}

\section{Discussion}
\subsection{Parralel vs Sequential Stacking}

\begin{figure}[ht]
\vskip 0.2in
\begin{center}
\centerline{\includegraphics[width=\columnwidth]{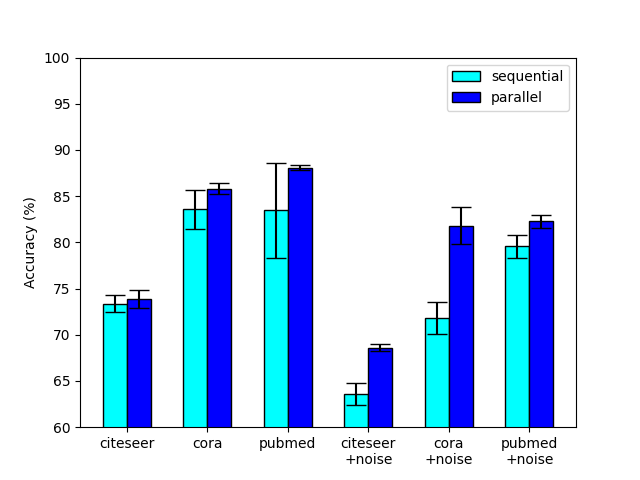}}
\caption{Comparison of sequential vs. parallel layers stacking}
\label{fig:sequential_parallel}
\end{center}
\vskip -0.2in
\end{figure}

Compared to the traditional sequential stacking of GNN layers, NODE-SELECT adopts the approach of stacking its layers in parallel. Based on our experiments, we have found that the parallel stacking of these selective layers yielded better and more stable results. Figure \ref{fig:sequential_parallel} illustrates a comparative study of these two stacking options. As demonstrated, the parallel setting proves to be more beneficial with its results reaching higher accuracy and lower variance. The sequential layout forces a layer $l$ to depend on the set of selected nodes $\mathcal{V}_{s}^{(l-1)}$ of a previous layer $l-1$. In the rare cases that the composition of $\mathcal{V}_{s}^{(l-1)}$ is not completely suitable to the weights of layer $l$, the model may result in a much lower performance; thus observing a higher variance and reduced accuracy. The parallel layout removes this dependence issue by stacking its layers side-by-side and having them learn independently as in the ensemble method \cite{dietterich2000ensemble}. 

\subsection{Layers Operation}
\begin{figure}%[ht]
\vskip 0.2in
\begin{center}
\centerline{\includegraphics[width=\columnwidth]{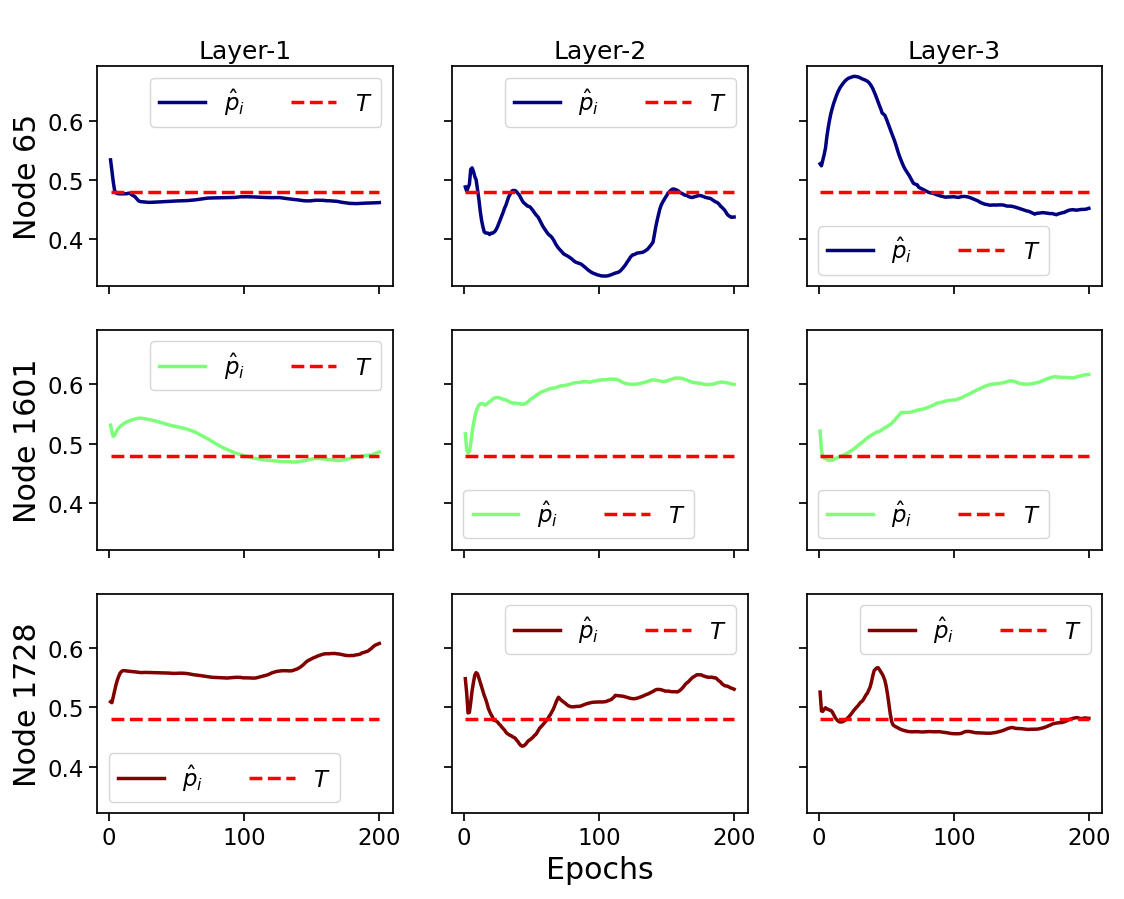}}
\caption{Learning plot of $\hat{p}_i$ during a NODE-SELECT model training. Red dashed line {\color{red} - - -} represents the threshold parameter $T$. For each node (65, 1601, 1728), an independent layer learns learns its \textit{selection} (allowing it to share its embedding information). A given node $v_i$ is selected when its $\hat{p}_i$ signal is at least as high as the threshold $T$.}
\label{fig:p_hat_i}
\end{center}
\vskip -0.2in
\end{figure}

\begin{figure}%[ht]
\vskip 0.2in
\begin{center}
\centerline{\includegraphics[width=\columnwidth]{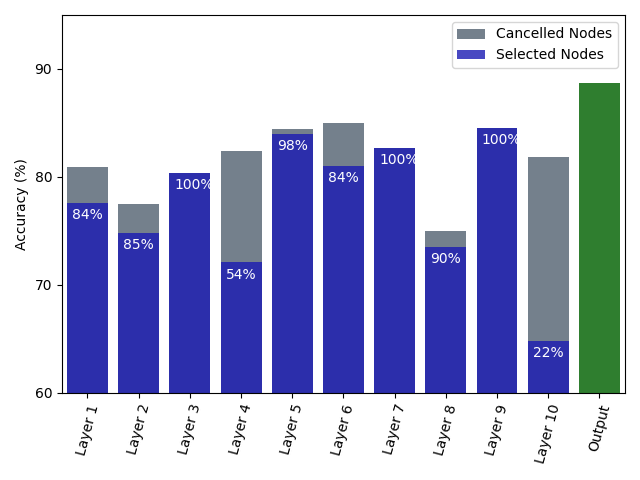}}
\caption{Accuracy score by layer and output for model trained on Pubmed. Blue color displays the percentage of selected nodes $\mathcal{V}_{s}^{(l)}$. The green color displays the reported accuracy of the final output, obtained by summing the embeddings from the 10 layers.}
\label{fig:layers_Vstar_comparison}
\end{center}
\vskip -0.2in
\end{figure}

Because of the parallel configuration, any given NODE-SELECT layer operates independently. Each layer separately makes a node selection that yields to a node embedding. Figures ~\ref{fig:p_hat_i} and ~\ref{fig:layers_Vstar_comparison} contrast these layers' differences
in terms of their predicted sensitivity value, accuracy score, and proportion of selected nodes for experiments on the Cora and Pubmed Dataset. In Figure ~\ref{fig:p_hat_i}, the $\hat{p}_i$ values of 3 nodes from the training set are predicted by 3 parallel layers on a Cora experiment. Because of the layers' independence, a node's $\hat{p}_i$ values from distinct layers are also independent. The independence is illustrated with Layer-1 learning to output high values above the threshold for node 1728 but conservative values node 1601, yet Layer-3 does the opposite. In Figure ~\ref{fig:layers_Vstar_comparison}, the accuracy results ranging from 75 to 85\% for the 10 layers used in a trained model are displayed with the selection percentage ranging between 22 and 100 \% (in blue). For instance, layer 6 reached the highest accuracy of 85\% with a selection percentage of 84\% while layer 3  had the third lowest accuracy score of 80\% with 100\% selection proportion. As demonstrated in the figure, a layer's accuracy performance is not correlated to its proportion in size of selected nodes. However, based on our experiments, we found that a layer that more effectively \textit{filters} the most noise-propagating nodes is more likely to reach a higher accuracy. We also see the reported accuracy (in green) of the final output, obtained by summing the embeddings of the 10 layers, being much higher than any layer's individual accuracy score.

\subsection{Scalability}

\begin{figure}%[ht]
\vskip 0.2in
\begin{center}
\centerline{\includegraphics[width=\columnwidth]{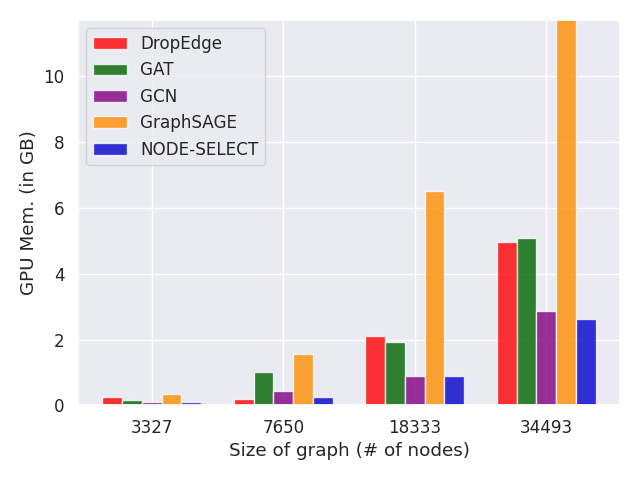}}
\caption{Scalability of NODE-SELECT compared to other methods when graph size increases.}
\label{fig:scaling-image}
\end{center}
\vskip -0.2in
\end{figure}

A direct benefit of stacking our layers in parallel is that our method is very scalable. Because our method is composed of ensembled 1-layer GNN, its memory usage is very effective. For example, the number of trainable parameters can be estimated with $L \left (F' ( 3 + F)\right )$; with $L$ describing the number of layers, $F'$ the output dimension, and $F$ the input dimension. Figure \ref{fig:scaling-image} contrasts how NODE-SELECT scales with larger graphs when compared to some baseline frameworks. As demonstrated,  NODE-SELECT scales to larger graphs comparably to GCN and much better than GAT, GraphSAGE, and DropEdge. As the graph gets larger, the amount of memory required for the learning to take place also increases and the challenge of being scalable affects many current GNN \cite{zhou2018graph,wu2020comprehensive}. Nonetheless, NODE-SELECT adapts very well to larger graphs.

\subsection{Effect of Parameters $L$ and  $T$}
In contrast to other GNNs, NODE-SELECT only has two configurable parameters that affect the model's performance. The parameter $L$  guides the model's fitting behavior while $T$ guides its selection mechanism. In our study, we found that any arbitrary number of layer leads to the a good performance. However, depending on the size and properties of the graph, too few layers may cause the model to under-fit while too many to over-fit. Table \ref{tab:table_effect_L} provides a simple illustration of the effect of increasing the parameter $L$. Using only 1 layer causes the model to under-fit with an accuracy of 94\%. Using 20 or more (100) layers causes the model to over-fit with accuracy scores that are below the well-fit models trained with 5 or 10 layers. Particularly, as a NODE-SELECT model uses more layers (past the optimal number), each individual layer becomes weaker. This decrease in performance results from the fact that each layer learns the patterns pertaining to a specific region in a graph and thus generalizes poorly.

\begin{figure}%[ht]
\vskip 0.2in
\begin{center}
\centerline{\includegraphics[width=\columnwidth]{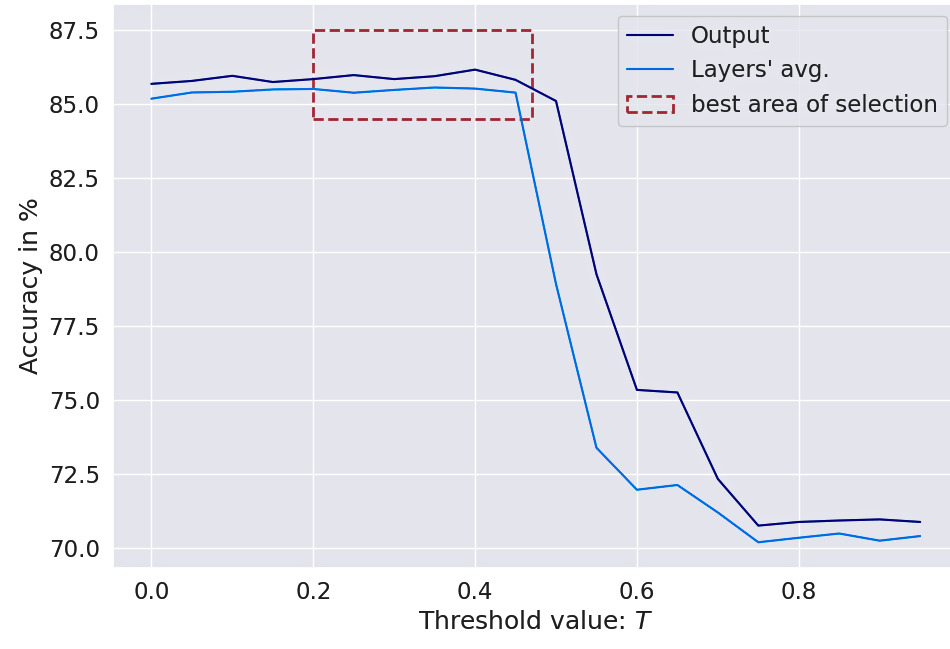}}
\caption{Results of using variable value of the threshold parameter $T$ from Cora experiments. Red rectangle contours area with high accuracy where selection is most \textit{diverse} (great variance of selection).}
\label{fig:AccxV}
\end{center}
\vskip -0.2in
\end{figure}

A NODE-SELECT model will adjust its weights so as to retain as much embedding information as possible during training. Therefore, a NODE-SELECT only begins to cancel nodes when the threshold parameter is not too small; for instance above $T$ value of $0.3$. However, using low values for the threshold results in performance that is only marginally lower than with moderate threshold. Also, the selection mechanism is a lot more effective when the graph is not small. In our experiments with smaller graphs (Cora and Citeseer), an effective layer only cancelled a minimal number of nodes (i.e. between 0 and 10\%). Figure \ref{fig:AccxV} displays the effect of changing $T$ on the model accuracy. Using the parameter $T$ allows the model to conservatively remove subset of nodes that are not needed to lower its loss. Using a $T$ value in a range of $0.3\le T \le 0.49$ gives the best results in terms of both accuracies and node cancellations. However the application of a large $T$ value leads the model to cancel too many nodes, thereby losing crucial information from potentially important nodes.

\begin{table}%[ht] 
\caption{Results of using variable number of layers. Rows in blue are for under-fit models and rows in red for over-fit models.}
\label{tab:table_effect_L}
\vskip 0.15in
\begin{center}
\begin{tabular}{|c|c|c|c|}
\hline
\multirow{2}{*}{\shortstack{\# of \\ Layers}} & \multicolumn{3}{|c|}{Co-CS} \\\cline{2-4}
& \shortstack{Layers mean\\ accuracy} & \shortstack{Model \\ accuracy}& \shortstack{Layers avg. size \\of selection} \\\hline
\cellcolor{LightCyan}1    &\cellcolor{LightCyan} 94.0&\cellcolor{LightCyan}94.0 &\cellcolor{LightCyan} 78\%\\\hline
5    &93.8&95.0& 51\%  \\\hline
10   &93.0&94.8&52\%  \\\hline
\cellcolor{IndianRed}20   &\cellcolor{IndianRed}85.5&\cellcolor{IndianRed} 94.8  &\cellcolor{IndianRed}64\%  \\\hline
\cellcolor{IndianRed}100  &\cellcolor{IndianRed}62.3&\cellcolor{IndianRed} 93.5 &\cellcolor{IndianRed} 66\%  \\\hline
\end{tabular}
\end{center}
\vskip -0.1in
\end{table}

\section{CONCLUSION}
We introduced NSGNN, a novel graph neural network for node-classification, which learns node embeddings by summing correlated embeddings learned by its layers. Inspired by the functioning of real-world graphs, our NODE-SELECT addresses the conceptual limitation of selective propagation based on the nodes global importance. As opposed to other frameworks which sequentially convolve the embeddings, thus removing key information in the embeddings, our NODE-SELECT relies on various complementary convolutions to enhance those key information.  Besides the reaching state-of-the-art performance in experiments which introduced noise propagation, its scalability to larger graphs is much more effective than other baselines. With a simple selection mechanism that allows our model to effectively adapt to the problem of noise-propagating nodes, we expect that our proposed method can adapt to real world problems where such mechanism can be very important such as in Botnet detection or cancellation of particular instances within a graph. Further research may also be done to additionally improve our method's performance by testing other ways to combine the independent layers' embedding or how other ways to ensemble the separate layers (i.e. boosting).

%%%%%%%%%%%%%%   Bibliography   %%%%%%%%%%%%%%
\normalsize
\bibliography{references}

\twocolumn[ % Method A for two-column formatting
  \begin{@twocolumnfalse} % Method A for two-column formatting
    \end{@twocolumnfalse} % Method A for two-column formatting
] % Method A for two-column formatting
\section{Appendix}
\label{appendix_NS}

\begin{table*}[ht]  
\caption{ The hyperparameters providing best accuracy for each baseline model on all datasets. These parameters are listed as (\# of layers / \# of neurons  used in hidden layers / learning-rate / optimizer's weight-decay/*additional-details). The same parameters were used in Node2vec whose additional details include (walk-length:20, context-size:10,walk-per-node:1,negative-sample=5)}
\label{tab:table_parameters}
\begin{center}
\begin{tabular}{|c|c|c|c|}
\hline
\textbf{Framework} & \textbf{Dataset} &\textbf{Acc.} &  \textbf{Configuration}\\
\hline
\multirow{8}{*}{FastGCN} & CiteSeer &74.0$\pm$1.0& 2 / 16 / 0.001 / 0.0005 / --- \\ \cline{2-4} & Cora &82.1$\pm$2.6 &  2 / 64 / 0.01 / 0.0005 / --- \\ \cline{2-4}&  PubMed  &87.6 $\pm$0.5 &  2 / 16 / 0.005 / 0.0005 / ---  \\\cline{2-4} &  Co-P &95.5$\pm$0.3 &  3 / 64 / 0.005 / 0.0005 / --- \\\cline{2-4}  & Co-CS &92.2 $\pm$0.4 & 3 / 128 / 0.005 / 0.0005 / ---\\\cline{2-4} &  Cora Full &60.8 $\pm$1.0 &3 / 128 / 0.005 / 0.0005 / ---    \\\cline{2-4} &  Amz-C &83.5 $\pm$2.2 &3 / 128 / 0.005 / 0.0005 / ---    \\\cline{2-4} &  Amz-P   &91.0$\pm$0.9 &3 / 128 / 0.005 / 0.0005 / ---     \\
\hline
\multirow{8}{*}{GAT} &CiteSeer &74.0$\pm$0.7&  2 / 64 / 0.0005 / 0.005 / attention-heads:8  \\ \cline{2-4} & Cora  &86.0$\pm$0.7 & 2 / 128 / 0.0005 / 0.005 / attention-heads:8 \\ \cline{2-4}&  PubMed  &86.4 $\pm$0.3 &3 / 64 / 0.01 / 0.00005 / attention-heads:8  \\\cline{2-4} &  Co-P &95.7 $\pm$0.1 &3 / 64 / 0.01 / 0.00005 / attention-heads:8  \\\cline{2-4}  & Co-CS &92.2 $\pm$0.2 &3 / 64 / 0.01 / 0.00005 / attention-heads:8 \\\cline{2-4} &  Cora Fullll &64.8 $\pm$0.5 &2 / 128 / 0.005 / 0.00005 / attention-heads:8    \\\cline{2-4} &  Amz-P & 93.7$\pm$0.6&2 / 128 / 0.005 / 0.00005 / attention-heads:8  \\\cline{2-4} &  Amz-C   &90.0$\pm$0.7 &2 / 128 / 0.005 / 0.00005 / attention-heads:8 \\
% &CiteSeer & Cora & PubMed &Co-P&Co-CS& Amz-P&Amz-C   & \\
\hline
\multirow{8}{*}{GCN} & CiteSeer &74.0$\pm$0.6&  2 / 128 / 0.0005 / 0.05 / ---  \\ \cline{2-4} & Cora  &85.0$\pm$0.7 &2/ 128 / 0.01 / 0.0005 / ---  \\ \cline{2-4}&  PubMed  &87.2 $\pm$ 0.2 &2 / 128 / 0.01 / 0.0005 / --- \\\cline{2-4} &  Co-P &95.9 $\pm$0.1 &2 / 64 / 0.01 / 0.0005 / ---  \\\cline{2-4} & Co-CS &93.1$\pm$0.2 &2 / 128 / 0.01 / 0.0005 / ---\\\cline{2-4} &  Cora Full &67.3 $\pm$0.5 &2 / 128 / 0.01 / 0.0005 / ---      \\\cline{2-4} &  Amz-P &93.5$\pm$0.2 &2 / 128 / 0.01 / 0.0005 / ---     \\\cline{2-4} &  Amz-C   &89.4$\pm$0.5 &2 / 128 / 0.01 / 0.0005 / --- \\
\hline
\multirow{8}{*}{GraphSAGE} & CiteSeer &73.7 $\pm$0.7&2 / 64 / 0.0005 / 0.005 / ---     \\ \cline{2-4} & Cora  &86.0$\pm$0.7 &2 / 64 / 0.0005 / 0.005 / ---    \\ \cline{2-4}&  PubMed  &86.2 $\pm$0.3 &2 / 64 / 0.05 / 0.0005 / ---  \\\cline{2-4} &  Co-P &95.4 $\pm$0.2 &2 / 64 / 0.005 / 0.0005 / ---   \\\cline{2-4}  & Co-CS &93.4$\pm$0.2 &2 / 64 / 0.001 / 0.0005 / ---     \\\cline{2-4} &  Cora Full &64.9 $\pm$0.3 &3 / 128 / 0.005 / 0.0005 / ---     \\\cline{2-4} &  Amz-P &94.4 $\pm$0.5 &  2 / 64 / 0.005 / 0.0005 / ---  \\\cline{2-4} &  Amz-C   &90.2 $\pm$0.5 & 3 / 128 / 0.005 / 0.0005 / --- \\
\hline
Node2vec & * & *   & 1 / 64 / 0.005 / --- /--- \\ 
\hline
\multirow{7}{*}{NODE-SELECT} & 
CiteSeer &74.1$\pm$1.1& 3 / --- / 0.01 / 0.05 / depth:2 \\ \cline{2-4} & Cora  & 85.8$\pm$0.6 & 3 / --- / 0.005 / 0.05 / depth:2 \\ \cline{2-4}&  PubMed  & 88.1 $\pm$0.3 &  8 / --- / 0.05 / 0.00005 / depth:1  \\\cline{2-4} &  Co-P &96.5$\pm$0.1 &10 / --- / 0.005 / 0.00005 / depth:1  \\\cline{2-4}  & Co-CS &94.8$\pm$0.1 &8 / --- / 0.005 / 0.00005 / depth:1  \\\cline{2-4} &  Cora Full &67.3 $\pm$0.6 &
8 / --- / 0.005 / 0.00005 / depth:1   \\\cline{2-4} &  Amz-C   &89.6$\pm$0.4 &25 / --- / 0.001 / 0.00005 / depth:1\\\cline{2-4} &  Amz-P   &94.4$\pm$0.3 &25 / --- / 0.001 / 0.05 / depth:1 \\
\hline
\end{tabular}
\end{center}
\end{table*}
\subsection{Effect of Number of layers}
\begin{figure}[ht]
  \centering
  \includegraphics[width=0.8\linewidth]{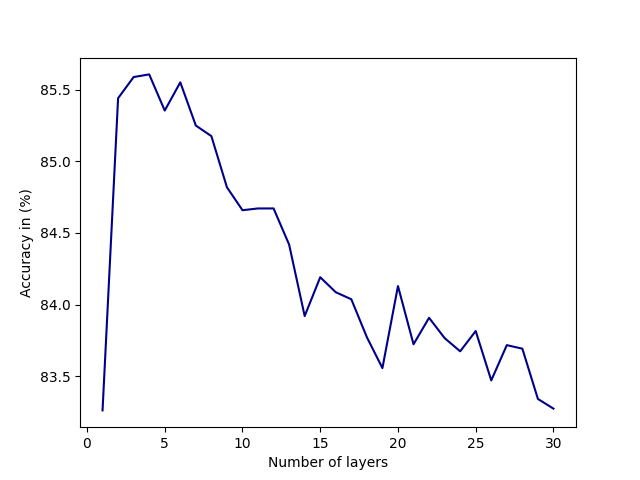}
  \caption{Number of layers x Avg. Accuracy}
  \label{fig:avg_acc_layers}
\end{figure}

\begin{figure}[ht]
  \centering
  \includegraphics[width=0.8\linewidth]{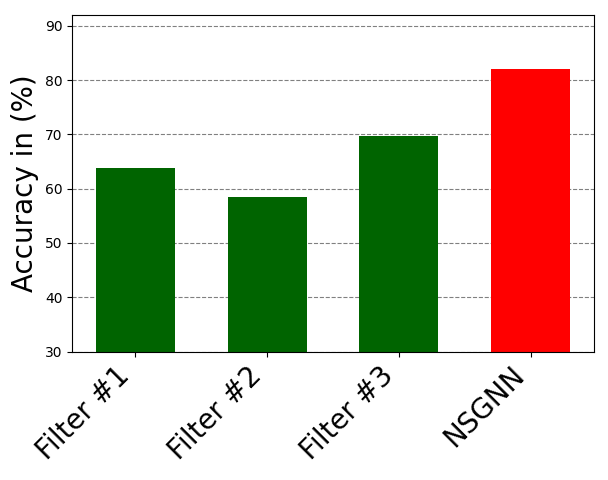}
  \caption{Using less than optimal number of layers in NODE-SELECT model on Amz-C experiments. Model under-fit reaching accuracy of 82\%}
  \label{fig:amz_layers_3}
\end{figure}

\begin{figure}[ht]
  \centering
  \includegraphics[width=0.8\linewidth]{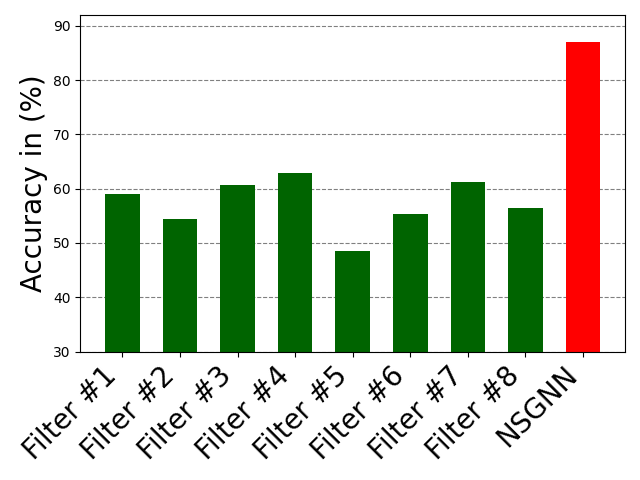}
  \caption{Increasing number of layers in NODE-SELCT model on Amz-C experiments. Model's performance increases in terms of accuracy gaining 5\% points, reaching score of 87\%.}
  \label{fig:amz_layers_8}
\end{figure}

There is an optimal number of layers that should be included in a NODE-SELECT model. Exceeding this number of layers increases the likelihood the model will over-fit while using less causes it to underfit. Figure ~\ref{fig:avg_acc_layers} displays the impact of surpassing the optimal number of layers (3-5) in the Cora dataset. At the optimal number of layers, the model's performance (average accuracy for 10 random splits) is at its peak. However, the increase in the number of layers prior to reaching that optimal number increases the performance of the NSGNN model. As more layers are added, NODE-SELECT captures the relationship patterns more efficiently and thus its predictive performance improves. The latter increase in performance is due to the specialization of the layers, which together complement each the other's inaccuracies. Figure ~\ref{fig:amz_layers_8} illustrates the effects of the addition of layers (in green) on the overall model (in red) on the Amazon-Computers dataset. In Figure ~\ref{fig:amz_layers_3}, the model's accuracy reaches an accuracy score of about 82\% while best layers obtains accuracy at exactly 70\%. As 5 more layers are added in the illustration of figure ~\ref{fig:amz_layers_8}, the highest accuracy reached by any layers drops 62\% while the NSGNN accuracy improves to 87\%. 
 
\subsection{Comparison of Layers' embeddings}
\begin{figure}[ht]
 \vskip 0.2in
  \centering
  \includegraphics[width=0.8\linewidth]{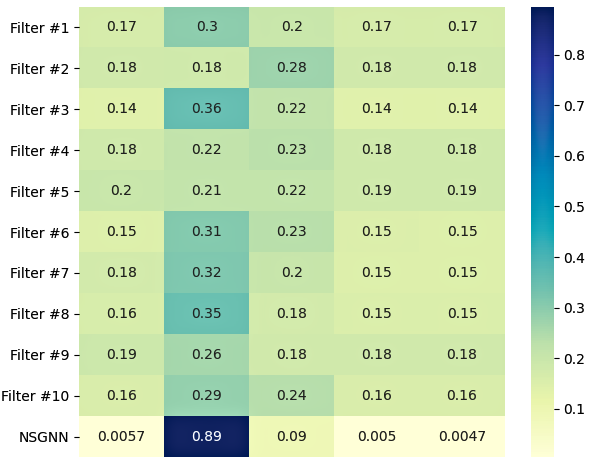}
  \caption{Increasing number of layers in NODE-SELCT model on Amz-C experiments. Model's performance increases in terms of accuracy gaining 5\% points, reaching score of 87\%.}
  \label{fig:embed_green_cop}
%   \vskip -0.2in
\end{figure}

Compared to the traditional sequential stacking of layers, NODE-SELECT adopts the well-known ensemble method for which each layer independently learns the nodes embedding based on a primary selection of propagating nodes. In our study, we observed that the average cosine-similarity of these embeddings per node is always high ($\geq0.85$). Henceforth, by combining these sets of independent yet correlated embeddings, the model subsequently learns a final embedding that is more precise. Figure ~\ref{fig:embed_green_cop} resulting embeddings from 10 independent layers for the first node on the Co-P experiment. Noticeably, the majority of the embeddings are very similar. Based on our experiments, we deduced that not all of the layers' embeddings need to match for a good node prediction. As long as there is a general harmony in a sufficient number of layers, the model will depend on these more frequent harmonious embeddings to form its final embedding. 

\subsection{Complex NODE-SELECT Layer}
\begin{figure}[ht]
 \vskip 0.2in
  \centering
  \includegraphics[width=0.8\linewidth]{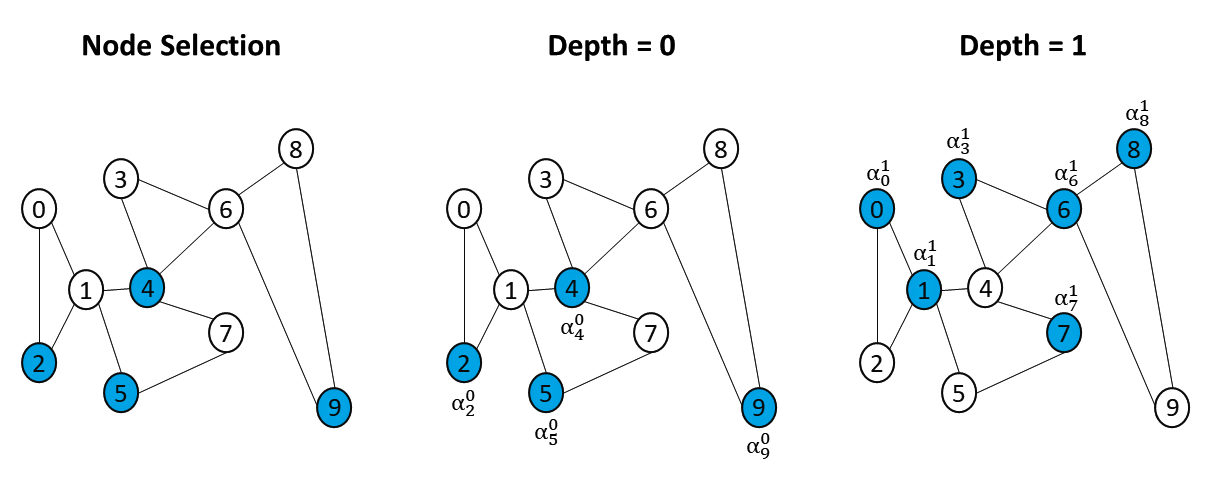}
  \caption{Sequential message propagation in layer. First, the layer selects a subset of propagating nodes based on their global importance. Those selected vertices are represented in blue: $\mathcal{V}_{S}^{l}=\{2, 4, 5, 9\}$. At depth $q=0$, only the selected nodes $\{2, 4, 5, 9\}$ are allowed to share a proportion $\alpha_{(\cdot)}^{(0)}$ of their embedding. At depth $q=1$, only the 1-hop neighbors of the initially selected nodes are allowed to share another proportion $\alpha_{(\cdot)}^{(1)}$ of their updated contribute to the message-passing.}
  \label{fig:layer_NS}
  \vskip -0.2in
\end{figure}

Beyond the simple selection mechanism that we presented in our paper, we also implemented a sequence mechanism that is more appropriate for smaller graphs. Mainly, this additional sequence mechanism allows each selected node to propagate information up to its $q$-hop neighbours. Figure ~\ref{fig:layer_NS} provides an illustration of this sequence mechanism. 

Starting with the subset of propagating nodes, a layer sequentially learns a global weight coefficient to perform the message-passing operation across $Q$-hop neighborhoods. For each $q = 0,1,..,Q\text{-}1$ depths, a corresponding weight coefficient $\alpha_{i}^{(q)}$ is calculated:
\begin{equation}
\alpha_{i}^{(q)} = \sigma(\textbf{W}_{2}\,(y_i^{(q)}\parallel q^{*}))
    \end{equation}
, in which $\textbf{W}_{2} \in \mathbb{R}^{1 \times (F+Q)}$ denotes a learnable matrix,  $y_i^{(q)}$ the $q$th updated feature vector of $v_i$, and $q^{*}$ the one-hot encoded vector of the depth $q$. 
 Upon learning this selection-depth adapted coefficient, a node's feature updates as:

\begin{equation}
y_i^{(q+1)} = y_i^{(q)} + \sum_{j \in \mathcal{N}_{(i)}}({\alpha_{j}^{(q)}\cdot y_j^{(q)}})  
\end{equation}

. Lastly, we implement a noise layering mechanism so as to regulate the amount of noise information that is being learned though the message-passing operations. Our motivation for the latter mechanism comes from the assumption that there exists a minority of nodes that do not need to aggregate their neighbors' information. We adapt our updating operation so that the layer tries to maintain an  appropriate balance between learned neighboring information and each node's own feature. Therefore, after $Q$ updates, the layer then calculates a final global coefficient $c_i$ though the use of a matrix $\textbf{W}_{3} \in \mathbb{R}^{1 \times (2 F)}$, 

\begin{equation}
c_i= \sigma(\textbf{W}_{3}\,(y_i^{(Q-1)}\parallel y_i^{(q=0)})
    \end{equation}

, where $Q-1$ denotes the last depth. The $c_i$ coefficient benefits in adjusting the learning of a given a node such that the layer may layer potential noise acquired during the aggregation process. Hence, the final embedding output by a NSGNN layer $l$ can be formulated as:
\begin{equation}
h_i^{(l)}= (1-c_i)\cdot y_i^{(q=0)} + c_i \cdot y_i^{(Q-1)}  
    \end{equation}
% =========================

%%%%%%%%%%%%  Supplementary Figures  %%%%%%%%%%%%
%\clearpage

%%%%%%%%%%%%%%%%   End   %%%%%%%%%%%%%%%%
%\end{multicols}  % Method B for two-column formatting (doesn't play well with line numbers), comment out if using method A
\end{document}